%% file: main.tex
\documentclass[10pt,twocolumn,letterpaper]{article}

\usepackage[pagenumbers]{cvpr} 

\usepackage{graphicx}
\usepackage{amsmath}
\usepackage{amssymb}
\usepackage{booktabs}

\input{setup/package}

\input{setup/macros}

\input{setup/graphicspath}

\usepackage[pagebackref,breaklinks,colorlinks,bookmarks=false,citecolor=blue,linkcolor=blue]{hyperref}

\usepackage[capitalize]{cleveref}

\crefname{section}{Sec.}{Secs.}
\Crefname{section}{Section}{Sections}
\Crefname{table}{Table}{Tables}
\crefname{table}{Tab.}{Tabs.}

\begin{document}

\title{
Fine-grained Controllable Video Generation via Object Appearance and Context
}
\author{Hsin-Ping Huang$^{1,2}$ \hspace{3mm}
Yu-Chuan Su$^{1}$ \hspace{3mm}
Deqing Sun$^{1}$ \hspace{3mm}
Lu Jiang$^{1}$ \hspace{3mm}
\\
Xuhui Jia$^{1}$ \hspace{3mm}
Yukun Zhu$^{1}$ \hspace{3mm}
Ming-Hsuan Yang$^{1,2}$
\vspace{2mm}
\\
$^{1}$Google Research \hspace{3mm}
$^{2}$UC Merced
}

\twocolumn[{
\renewcommand\twocolumn[1][]{#1}
\maketitle
}
]

\begin{abstract}
    \input{sec0_abstract}
\end{abstract}

\input{sec1_intro}

\input{sec2_related}

\input{sec3_approach}

\input{sec4_results}

\input{sec5_conclusions}

\input{sec6_appendix}

{
\small
\bibliographystyle{ieee_fullname}
\bibliography{egbib}
}

\end{document}

%% file: setup/package.tex
\usepackage{multirow}
\usepackage{tabularx}
\usepackage{etoolbox,siunitx}
\robustify\bfseries
\usepackage{enumitem}
\usepackage{float}
\setlist[itemize]{noitemsep, topsep=0pt}

\usepackage{comment}

\usepackage[table]{xcolor}

\usepackage{adjustbox}

\usepackage[super]{nth}

%% file: setup/macros.tex
\newcommand{\ignore}[1]{}
\makeatletter
\newcommand{\printfnsymbol}[1]{%
        \textsuperscript{\@fnsymbol{#1}}%
}
\makeatother

\newcommand{\topic}[1]{\smallskip\noindent\textbf{#1}}

\newcommand{\ours}{FACTOR}

%% file: setup/graphicspath.tex
\graphicspath{{figure}, {images}, {example}}

%% file: sec0_abstract.tex
Text-to-video generation has shown promising results. However, by taking only natural languages as input, users often face difficulties in providing detailed information to precisely control the model’s output. 
In this work, we propose fine-grained controllable video generation (\textbf{FACTOR}) to achieve detailed control.
Specifically, FACTOR aims to control objects' appearances and context, including their location and category, in conjunction with the text prompt.
To achieve detailed control, we propose a unified framework to jointly inject control signals into the existing text-to-video model.
Our model consists of a joint encoder and adaptive cross-attention layers. 
By optimizing the encoder and the inserted layer, we adapt the model to generate videos that are aligned with both text prompts and fine-grained control.
Compared to existing methods relying on dense control signals such as edge maps, we provide a more intuitive and user-friendly interface to allow object-level fine-grained control.
Our method achieves controllability of object appearances without finetuning, which reduces the per-subject optimization efforts for the users.
Extensive experiments on standard benchmark datasets and user-provided inputs validate that our model obtains a 70\% improvement in controllability metrics over competitive baselines. Project page: \url{https://hhsinping.github.io/factor}

%% file: sec1_intro.tex
\begin{figure*}[t]
    \centering
    \includegraphics[width=\textwidth]{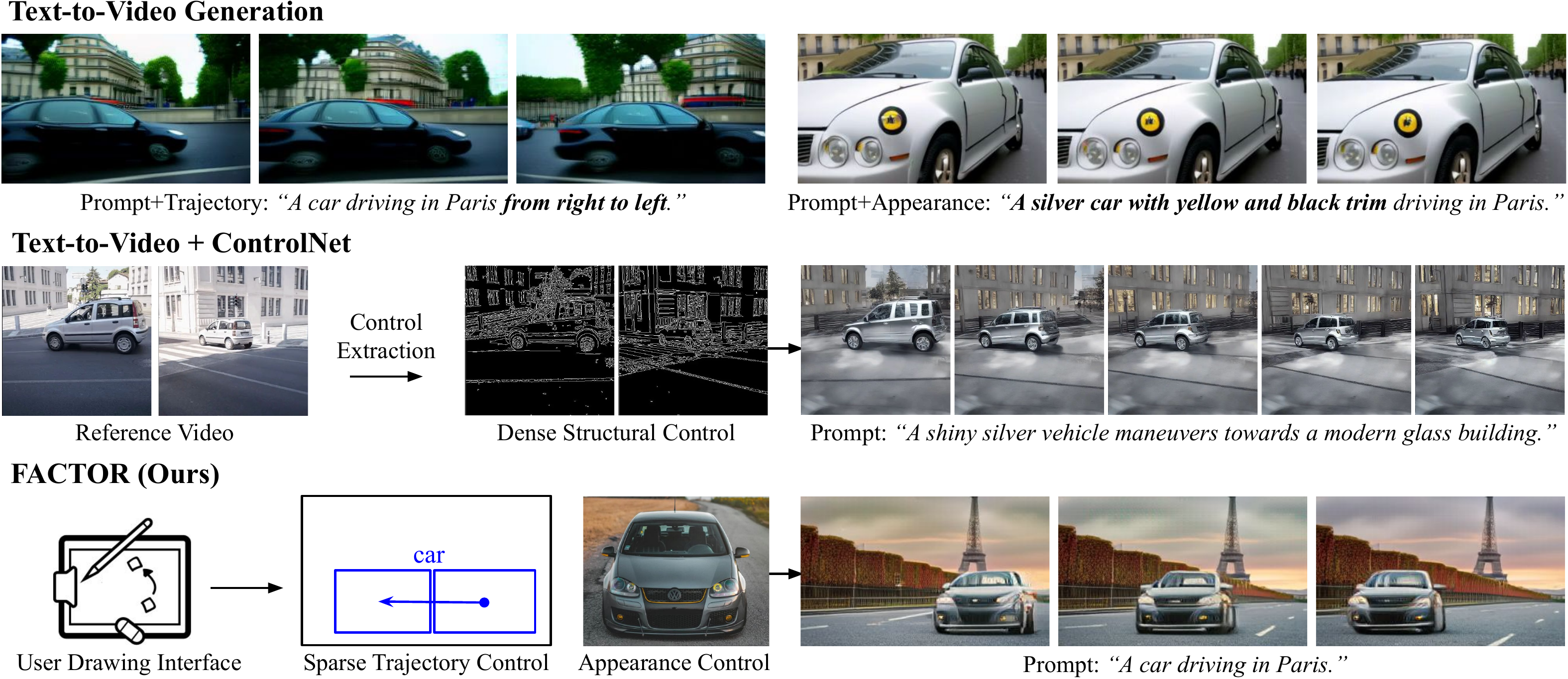}
    \captionof{figure}{\textbf{Text-to-video generation}~\cite{villegas2022phenaki} has limited controllable ability through user-provided prompts. %
    Even when the users augment the text prompt with additional description, the model has difficulty controlling the precise movement and appearance of the object.
    \textbf{Text-to-Video+ControlNet}~\cite{zhang2023controlvideo} achieves promising visual quality while requiring dense control signals extracted from a reference video.
    \textbf{FACTOR (Ours)} improves the controllability through user-friendly inputs to control the: 1) precise movement of subjects through hand-drawing trajectory and 2) visual appearance by providing reference examples. 
    }
    \label{fig:teaser}
\end{figure*}

\section{Introduction}
\label{sec:intro}

With numerous text-to-video models~\cite{hong2022cogvideo,wu2021nuwa,wu2021godiva,ho2022imagen,singer2023makeavideo,blattmann2023videoldm,Khachatryan2023text2video,wang2023modelscope,ge2023preserve,villegas2022phenaki,zhou2023magicvideo}  demonstrating impressive visual quality in their generated results,
users can now translate their creative ideas into video content with a simple text prompt and a tap.
However, it often requires careful design and iterative revision of the text prompt to generate a video that fulfills users' needs.
As shown in~\cref{fig:teaser} (top), to generate videos with specific object movements and appearances, the users can augment the text prompt with additional descriptions such as ``from right to left" and ``yellow and black trim." %
However, current text-to-video models might fail to generate videos that reflect the additional, fine-grained descriptions due to the high flexibility of text prompts and the limited capability of models, specifically in controllability.

Recently, a few approaches have been developed to enhance the controllability of text-to-video models by injecting additional inputs. Among this direction, structure, and appearance are the two prevalent and intuitive forms of user control.
Several models~\cite{wu2022tuneavideo,Khachatryan2023text2video,zhang2023controlvideo,chen2023controlavideo} are proposed to generate videos conditioned on structural controls such as optical flow, depth, and edge maps.
Similarly, video editing~\cite{vid2vid-zero,liu2023videop2p,qi2023fatezero,tokenflow2023} takes a video as input and uses natural languages to manipulate the content and style. 
Although these models have demonstrated promising visual results, they rely on control inputs typically dense in time and space, \ie, provided for each pixel in each frame.
Unlike text-to-image generation, providing such inputs for the whole video by manual hand drawing is impractical.
Thus, the structural inputs are usually obtained from a reference video (\cref{fig:teaser} middle).
In addition, a few attempts have been made to achieve appearance control, \ie, subject customization for video generation.
However, these methods either rely on taking the dense control as input~\cite{Khachatryan2023text2video,zhao2023makeaprotagonist,wu2022tuneavideo} or require finetuning per subject/domain using a set of images~\cite{ruiz2022dreambooth,guo2023animatediff,blattmann2023videoldm}.
Moreover, existing controllable video generation models are limited to taking only a single type of control, either structural~\cite{Khachatryan2023text2video} or appearance~\cite{blattmann2023videoldm}. To achieve controllability with multiple control signals, the model has to be finetuned in a sequential manner~\cite{wu2022tuneavideo}.

In this work, we tackle fine-grained controllable video generation via object appearance and context, a user-friendly controllable video generation framework.
In this framework, the users first specify the objects in the video and then provide fine-grained control to generate each object, including their appearance and context. The context to generate each entity includes 1) the description of the entity, 2) the user-drawing trajectory, and 3) the user-provided reference image to achieve both structural and appearance control.
This setting has the following advantages:
First, our model accepts sparse hand-drawing object trajectories as intuitive inputs that require minimal effort for users to provide, in contrast to dense structural guidance, such as edge maps.
Second, unlike prior works that utilize per-subject finetuning to achieve appearance control, our method is finetuning-free, and only an inference pass is needed for subject customization, reducing the users' efforts.

We propose a joint encoding and adaptive cross-attention framework to facilitate fine-grained, controllable video generation. We develop the method on top of an off-the-shelf text-to-video model~\cite{villegas2022phenaki}. 
Our controllable module is a unified framework for both types of control signals, and the module is learned by a single adapting process in contrast to prior works that finetune the models for multiple control sequentially~\cite{wu2022tuneavideo}. It is generic and can be extended to other types of control easily.
Specifically, all control signals are encoded into a single control sequence through a joint encoder, and the adaptive cross-attention layers are inserted into transformer blocks of the model to take the fine-grained control inputs.
During training, the newly inserted layers are updated while other layers of the pre-trained model are fixed.
This design preserves the text-to-video model's capability to generate high-quality videos while introducing adaptive layers that augment the model's ability to produce videos that satisfy the object-level fined-grained control.

We conduct quantitative studies on the standard benchmark dataset MSR-VTT~\cite{xu2016msrvtt} and perform a user study to validate the effectiveness of the proposed approach. 
Our method showcases a substantial 67\% and 73\% enhancement in absolute quality and trajectory control compared to the base text-to-video model, as observed through objective metrics.
The qualitative results show that our method achieves fine-grained, controllable video generation for object appearance and context.
We summarize our contributions as follows:
\begin{itemize}[leftmargin=*]
\setlength{\itemsep}{1pt}
\setlength{\parskip}{1pt}
\item We target the new form of \textit{fine-grained controllable video generation} task that aims to synthesize videos via appearance and context from easy-to-give user inputs. %
\item We propose a generic and unified framework for controllable video generation. It is achieved through adaptive training to augment text-to-video models for fine-grained control without test time optimization. 
\item We validate that our method offers better controllability compared to prior works and show the additional benefit of our model in creating complex interactions, which is challenging for existing text-to-video models.
\end{itemize}

%% file: sec2_related.tex
\begin{figure*}[t]
    \centering
    \includegraphics[width=\textwidth]{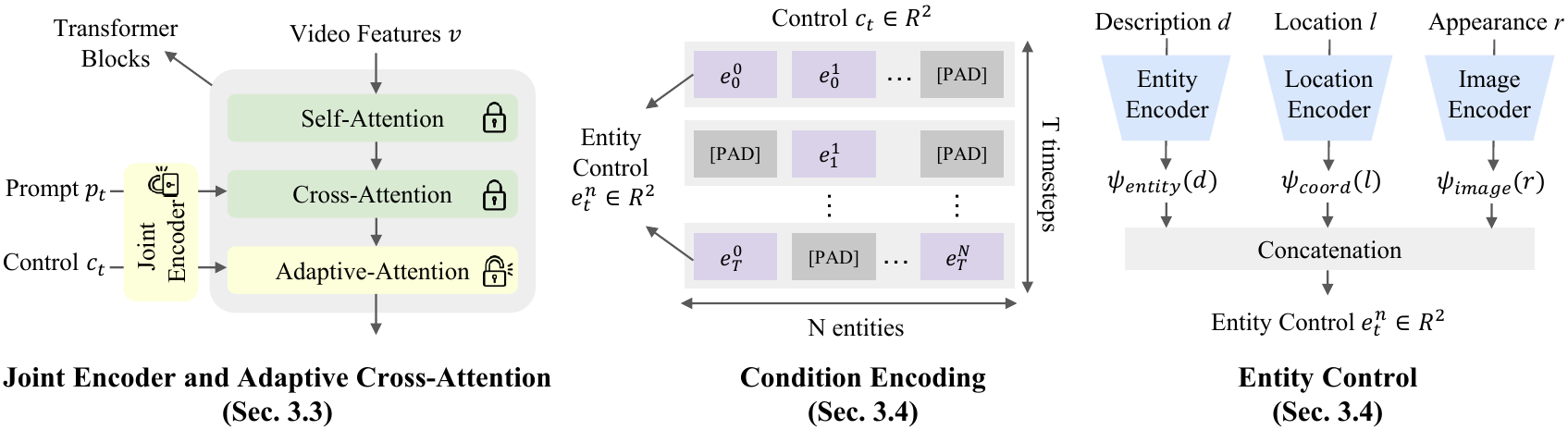}
    \vspace{-15pt}
    \captionof{figure}{\textbf{Overview.} %
    a) \textbf{Joint encoder and adaptive cross-attention}: 
    a joint encoder is learned to encode the prompt and control to capture the interaction between them.
    The adaptive layers are inserted into the transformer blocks of the text-to-video model to take new control signals.
    Only the inserted layers are optimized to adapt the model to generate videos satisfying the fine-grained control.
    b) \textbf{Condition encoding}:
    given T time steps, the embedding of control at time $t$, $c_t$, is formed by the control for N entities, $e^n_t$, where padding tokens replace the embedding of the non-existing entity.
    c) \textbf{Entity control}: the control for entity $n$ at time $t$, $e^n_t$, is formed by the embedding of the context including the description, location, and reference appearance of the objects. %
    }
    \label{fig:method}
    \vspace{-6pt}
\end{figure*}

\section{Related Work}
\label{sec:related}

\topic{Text-to-video generation.} 
Text-to-video models have successfully generated open-domain videos conditioned on text prompts by training on large-scale caption-video pairs.
Token-based models~\cite{wu2021godiva,wu2021nuwa,hong2022cogvideo} are first proposed for text-to-video generation, which utilizes an auto-regressive model to predict videos in the latent token space.
Several diffusion-based models~\cite{singer2023makeavideo,ho2022video,ho2022imagen} extend the 2D diffusion model to generate videos using 3D U-Net and a series of super-resolution modules.
Recent advances~\cite{Khachatryan2023text2video,wang2023modelscope,zhou2023magicvideo,blattmann2023videoldm,ge2023preserve} adopt a text-to-image diffusion model~\cite{rombach2021highresolution} to generate videos by inflating the model and incorporating the temporal layers.
While these models produce promising visual results, their controllability over generated videos is limited since text prompts cannot accurately convey precise control signals such as the location and appearance of objects.
In this work, we focus on improving the controllability of text-to-video models through user-friendly inputs.
We adopt a generative transformer model~\cite{villegas2022phenaki} as our base text-to-video model due to its fast inference speed and flexibility to take multiple control signals but the approach can be adapted to common diffusion-based models easily.

\topic{Controllable text-to-image generation.}
Various models have been proposed to improve the controllable ability over text prompts for text-to-image models, particularly the structure and appearance. 
ControlNet~\cite{zhang2023adding}, T2I-Adapter~\cite{mou2023t2i} and Uni-ControlNet~\cite{zhao2023uni} add various spatial control signals, including edges, depth, segmentation, human pose, to a pre-trained diffusion model.
GLIGEN~\cite{li2023gligen} adds spatial and appearance control to the text-to-image model.
Some works focus on structure-guided generation from semantic layouts~\cite{Avrahami_2023_CVPR,xue2023freestylenet,yang2023reco,zeng2022scenecomposer,ham2023modulating} and training-free approaches~\cite{bar2023multidiffusion,yu2023freedom,xie2023boxdiff,densediffusion,epstein2023selfguidance}. %

To control the appearance of subjects for text-to-image models, Textual Inversion~\cite{gal2022textual}, DreamBooth~\cite{ruiz2022dreambooth} and CustomDiffusion~\cite{kumari2022customdiffusion} either directly optimize a word vector to learn the target appearance, or finetune the text-to-image model on reference images of the subjects.
Several encoder-based methods~\cite{gal2023encoderbased,wei2023elite} are proposed to achieve customization without finetuning by projecting reference images of the subject into the word embeddings.
In this work, we study finetuning-free subject customization for videos with trajectory control.

\topic{Controllable text-to-video generation.}
To extend controllable text-to-image models to generate videos, 
several works~\cite{2023videocomposer,wu2022tuneavideo,Khachatryan2023text2video,ma2023follow,zhang2023controlvideo,xing2023make,chen2023controlavideo,liew2023magicedit,chen2023eve,chu2023medm,esser2023structure} adopt the ControlNet~\cite{zhang2023adding} backbone to condition the generation on a sequence of structural inputs including depth, pose, and edge maps.
By training a temporal attention layer built upon the ControlNet model, these methods learn to produce temporally consistent videos.
However, they require a dense sequence of structural inputs as the condition, which usually need to be extracted from reference videos. %
Similarly, video editing approaches~\cite{vid2vid-zero,yang2023rerender,chai2023stablevideo,shin2023editavideo,liu2023videop2p,qi2023fatezero,hu2023videocontrolnet,wang2023genlvideo,tokenflow2023,xing2023simda,ouyang2023codef,huang2023styleavideo,qin2023instructvid2vid} focus on changing the appearance and style of foreground objects and background through text prompts. These methods also rely on the dense structure provided in the input video, which is infeasible to be 
drawn by the users.
In contrast, our work focuses on controllable text-to-video generation that takes sparse and user-friendly control signals, which remains less explored in this field.

As for appearance control, a few approaches~\cite{wu2022tuneavideo,Khachatryan2023text2video,he2023animate,zhao2023makeaprotagonist} show video customization results using finetuning approach while still relying on the dense structural input to generate videos. Others~\cite{guo2023animatediff,blattmann2023videoldm} perform finetuning-based customization without spatial control.
In contrast, we developed a finetuning-free method to achieve appearance control of multiple subjects, and the model conducts spatial control using sparse trajectory inputs simultaneously.

%% file: sec3_approach.tex
\section{Method}
\label{sec:method}
As our method is developed upon a text-to-video generative transformer, we first briefly review the base model (\cref{sec:phenaki}). %
Then, we give an overview of our fine-grained controllable video generation (\cref{sec:overview}).
Next, we introduce the proposed method, which consists of a joint encoder of learning the interaction between different controls and an adaptive cross-attention module to adapt the text-to-video model to achieve fine-grained control (\cref{sec:adaptive}).
Finally, we present the encoding process of our entity-level fine-grained control and the data preparation pipeline (\cref{sec:encoding}). %

\subsection{Preliminaries: Text-to-Video Generation}
\label{sec:phenaki}
This section presents our base text-to-video generation model~\cite{villegas2022phenaki}.
It consists of an encoder-decoder model that encodes the video into discrete video tokens and a bidirectional transformer model that predicts the video tokens conditioned on the embedding of text prompts. 
Specifically, the video is encoded and flattened into a long sequence as the input to the transformer.
At training time, the tokens are replaced with a special token \texttt{[MASK]}, and the transformer model is optimized to predict the tokens at \texttt{[MASK]} location conditioned on the text embedding. %
We minimize the negative log-likelihood of conditional video token prediction for the masked tokens $v_t, t \in M$, $M$ denotes the subset of video tokens that are masked, 
$v$ denotes the video tokens sequence, $v_M$ denotes the masked version of $v$.
The model is trained by:
\begin{equation}
    \mathcal{L} ={-}\mathop{\mathbb{E}}\limits_{ v \in \mathcal{D}}  \sum_{t\in M} \log p(v_{t}|v_M, p) 
\end{equation}
At inference time, all tokens are replaced with \texttt{[MASK]}, and the transformer model iteratively predicts the tokens conditioned on $p$.

The video token prediction model contains a series of transformer blocks. 
The video embeddings are first input to the self-attention. Then, a cross-attention layer takes video and text embeddings as input to condition the video token prediction on the text embeddings.
We use a token-based generative transformer model for its fast inference and inherent flexibility in modeling various control signals directly within the transformer architecture.

\subsection{Overview}
\label{sec:overview}
Given a text prompt $p$ and a fine-grained control $c$ as inputs, our model aims to generate a video to satisfy both input conditions. Specifically, the users provide fine-grained control $c$ by 1) describing the desired entities in the video, 2) drawing their trajectories, and 3) providing a reference appearance image for each entity.
This pipeline helps the users to create videos intuitively. 
Specifically, the input control signal is given at $T$ timesteps. Assume there are $N$ entities in the video, we define our control as $c=\{c_t\}_{t=1}^{T}$, $c_t\in\mathbb{R}^2$.
At a single timestep $t$, the control is formed by a sequence of entity control, \ie, the embeddings of the $N$ entities $c_t=\{e_t^n\}_{n=1}^{N}$, $e_t^n\in\mathbb{R}^2$. The embeddings encode the desired condition to generate the $n^{th}$ entity at time $t$.

\subsection{Joint Encoder and Adaptive Cross-Attention}
\label{sec:adaptive}
We explain the main components of our method in this section, including a joint encoder and an adaptive cross-attention module.
Compared to existing works (\eg,~\cite{zhang2023adding}) that use a separate encoding process for each input condition, we utilize a joint encoder to encode both the text prompt $p$ and the fine-grained control $c$ simultaneously within the same transformer.
This design facilitates the interaction between the text prompt and fine-grained control, captured into the contextualized embeddings via the joint encoder.
Specifically, we use simple embedding layers to transform the discrete tokens of prompt $p$ and control $c$ into embedding sequences.
The two sequences are then concatenated and input to self-attention layers to obtain the contextualized embeddings.
Finally, the embeddings are split back into prompt and control embeddings.%

The next task is to incorporate the fine-grained control into the base text-to-video transformer.
Given a sequence of control embeddings, we extend the existing text-to-video model by inserting an adaptive cross-attention layer in each transformer block to take additional control. See~\cref{fig:method}. %
During training, we freeze the weight of the self-attention and the cross-attention layers of the pre-trained model. 
We train the joint encoder and the newly inserted adaptive cross-attention layer from scratch to adapt the text-to-video model to generate videos aligned with text and the new fine-grained condition.
By fixing the pre-trained weights, the capability to generate high-quality videos is preserved, and the model learns to generate videos satisfying the fine-grained control by updating only a small portion ($\approx$ 23\%) of the parameters.

Though we use a transformer-based model, our adaptive cross-attention layers may be extended to other text-to-video models, \eg, diffusion-based models, as their architecture usually contains a series of similar blocks of convolution and attention layers~\cite{rombach2021highresolution,blattmann2023videoldm}. %
Likewise, this module can be easily extended to take other control signals. This is achieved by first transforming the control into tokens and inputting them to our joint encoder and adaptive cross-attention layers. %
Instead of using multiple condition-specific methods, our adaptive process is a unified approach to condition the model on multiple control signals in a single training pipeline.

\begin{figure*}[t]
    \vspace{-6pt}
    \centering
    \includegraphics[width=1.0\textwidth]{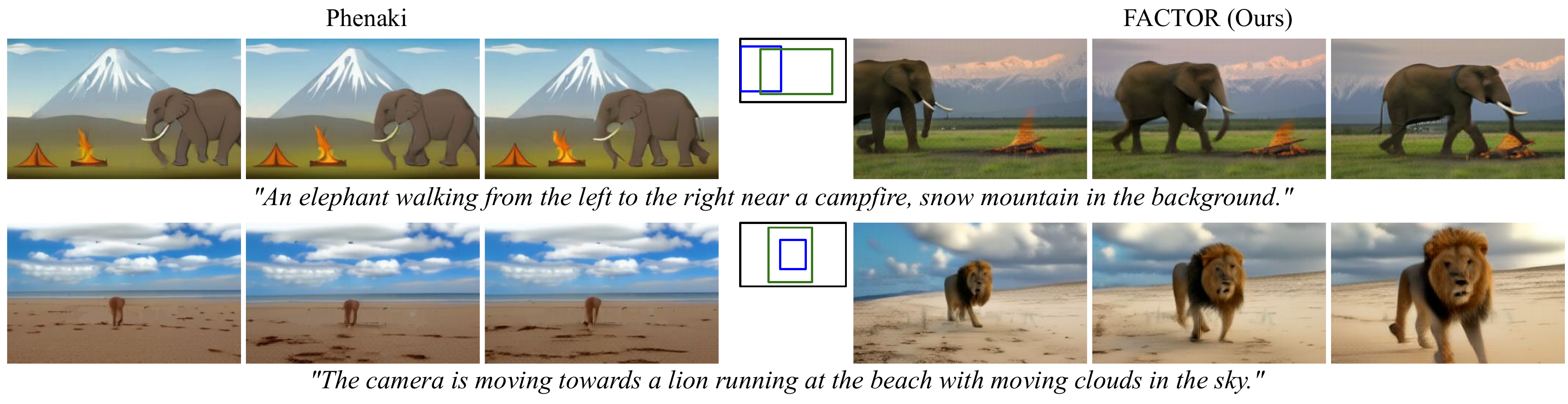}
    \vspace{-15pt}
    \caption{\textbf{Trajectory prompts.} To highlight that the prompt is not enough to achieve the fine control provided by our method, we augment the text prompt to describe the trajectories specified by the bounding boxes as inputs to the text-to-video model.
    Phenaki~\cite{villegas2022phenaki} fails to generate the correct object movement, while FACTOR successfully controls the movement of generated entities with our hand-drawing trajectory input.
    The blue and green boxes show the location of the object in the first and last frames, respectively. %
    } %
    \label{fig:direction}
    \vspace{-6pt}
\end{figure*}

\subsection{Condition Encoding and Entity Control}
\label{sec:encoding}

This subsection discusses our entity-level fine-grained control in detail.
The entity embeddings $e_t^n$ for generating the $n^{th}$ entity at timestep $t$ are constructed by encoding the context of each entity.
First, the description of the entity $d$ is given by text, \eg, a cat, and encoded by an embedding layer to transform the discrete tokens into embeddings $\psi_{entity}(d)$.
Second, the location of the entity $l$ is given by the top-left and bottom-right bounding box coordinates of the entity and encoded by an embedding layer as $\psi_{coord}(l)$.
Finally, the reference appearance $r$ of the entity is given by a \textit{single} example image and encoded by a CLIP image encoder with an MLP layer to reduce the feature size as $\psi_{image}(r)$.
The embeddings $e^n_t$ is the concatenation of description, location, and appearance embeddings of the entity:
\begin{gather}
e^n_t = Concat(\psi_{entity}(d^n_t), \psi_{coord}(l^n_t), \psi_{image}(r^n_t)).
\end{gather}
We replace the embeddings of $e^n_t$ with padding embeddings when the $n^{th}$ entity is missing at timestep $t$.
Though all the conditions can be thoroughly given at $T$ timesteps, we assume prompts, description, and appearance are fixed in the whole video, and only the location changes over time to resemble the user-friendly setup. 
In our user interface, the location is given by simply drawing a bounding box in the first frame and dragging it to move to the location in the last frame.

\label{sec:data}

In practice, very few video datasets contain the annotations of objects' trajectories and visual examples of the objects' appearance.
To train the model, we utilize an off-the-shelf object detector~\cite{raid} and tracking algorithm~\cite{Bewley2016_sort} to extract $N$ entities in the video clip and their locations at $T$ timesteps.
For each entity in the video, we need to collect a reference visual example as our appearance control $r$.
However, unlike image~\cite{ruiz2022dreambooth,kumari2022customdiffusion}, it is difficult to collect multiple videos of the same subject as training data.
We select the reference visual example of the subject from the same video.
To obtain reference images with more diverse appearances, we sample them from a longer video clip ($\approx$40 frames) outside our training timespan ($\approx$11 frames). We obtain the reference image of the subject by cropping the region using detected bounding boxes.
Finally, we use the collected triplet of text prompt $p$, control $c$, and video $v$ to train our model.

%% file: sec4_results.tex
\begin{figure*}[t]
    \vspace{-6pt}
    \centering
    \includegraphics[width=1.0\textwidth]{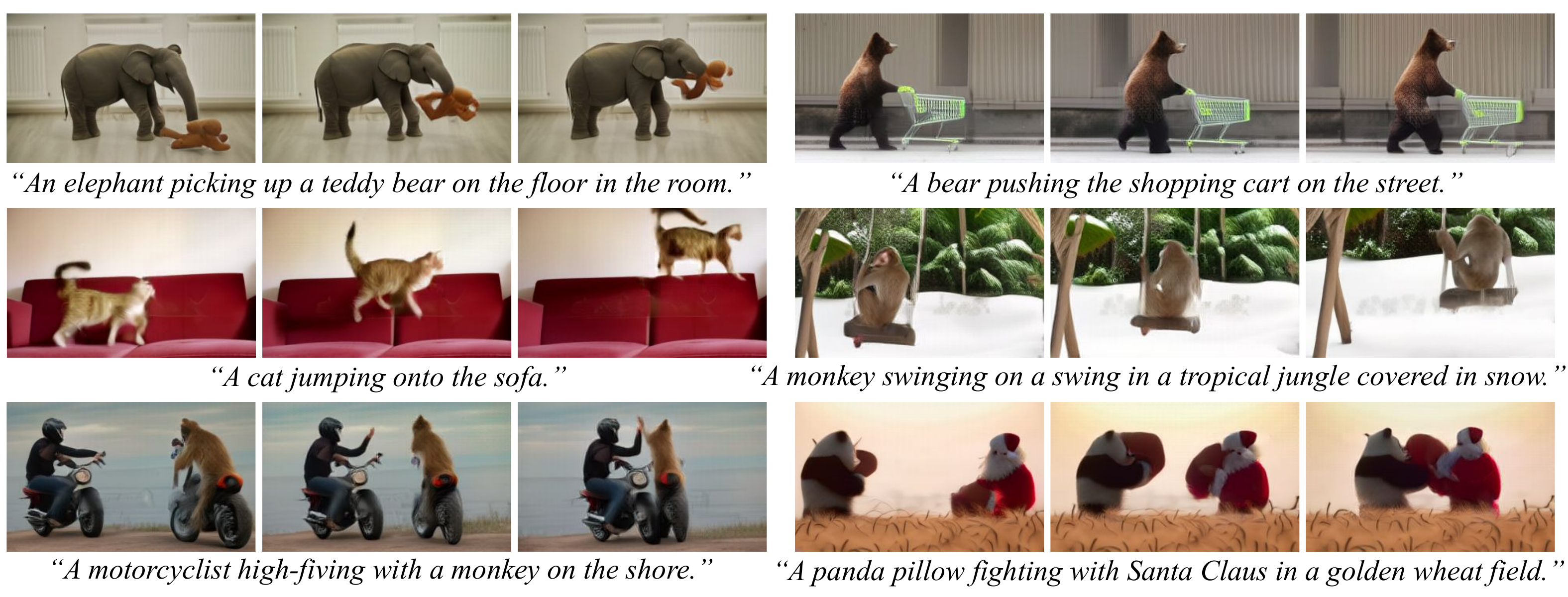}
    \vspace*{-18pt}
    \caption{\textbf{Trajectory control.} 
Given the trajectories of the two main entities in the videos as input, FACTOR brings an additional benefit to generate complex videos containing subject-object (top, middle) and subject-subject (bottom) interactions between two entities. The trajectory control inputs are omitted for simplicity. 
    }
    \label{fig:motion}
\end{figure*}

\begin{figure*}[t]
    \centering
    \includegraphics[width=1.0\textwidth]{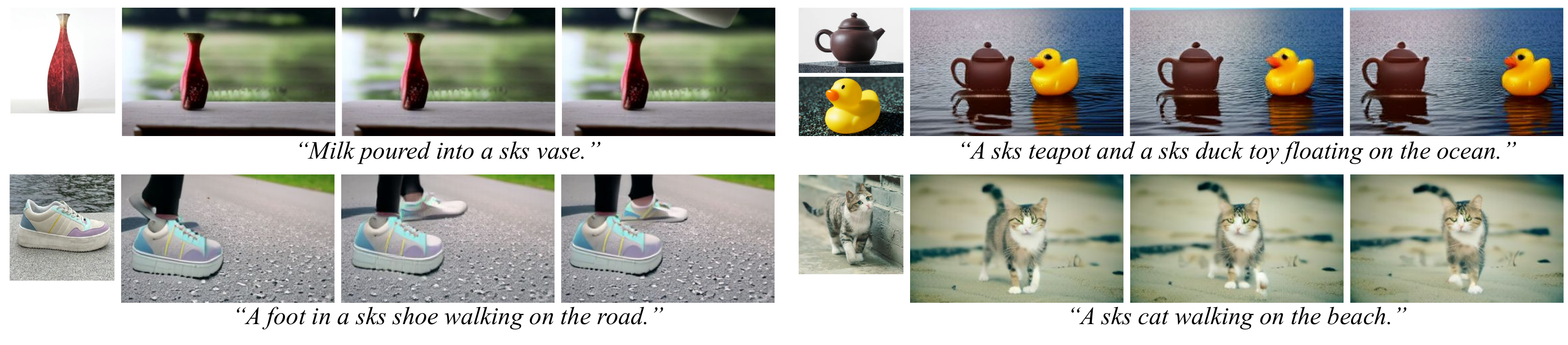}
    \vspace*{-18pt}
    \caption{\textbf{Appearance control.} FACTOR generates videos with the desired object appearances. The videos contain interaction for the customized subject (left), the composition of two customized subjects (top-right), and reasonable motion of live subjects (bottom-right). %
    }
    \label{fig:custom2}
    \vspace{-3pt}
\end{figure*}

\section{Experimental Results}
\label{sec:result}

\topic{Dataset.} 
We train our model on two datasets: WebVid~\cite{Bain21} and WebLI~\cite{chen2023pali}.
WebVid~\cite{Bain21} is a large-scale dataset of short videos and captions collected from stock footage sites. 
We use the 10M video-and-caption pairs in our training.
WebLI~\cite{chen2023pali} is a high-volume image-and-language dataset collected from the public web. 
We randomly sample a subset of WebLI with $\sim$500M image-and-text pairs in our training. 
We train our base text-to-video model from scratch using WebVid and WebLI data.
Then, we train our controllable video generation model using the same data with our augmented annotations $c$ (\cref{sec:data}).
We evaluate our model on MSR-VTT~\cite{xu2016msrvtt} dataset using zero-shot performance on the test set, which consists of 2,990 examples with 20 prompts per example.
Following~\cite{blattmann2023videoldm}, we generate one video per sample by randomly selecting one prompt.

\topic{Evaluation metrics.} 
We measure the video quality and the alignment between the generated video and different input conditions including text, trajectory, and reference images, using the following metrics:
1)~\textit{Fr\'echet Video Distance (FVD)} assesses the absolute video quality.
2)~\textit{CLIP text similarity (CLIP-T)} assesses the alignment between input prompts and generated video frames using CLIP embeddings.
3)~\textit{Average Precision (AP)} assesses the alignment between input object trajectories and the generated videos. We apply a pre-trained object detector~\cite{raid} to compare the detected bounding boxes between the generated frames and ground truth frames and report the AP score.
4)~\textit{CLIP video similarity (CLIP-V)} assesses the similarity between the reference images and object appearance in the generated video frames. We measure the CLIP embedding distance between the generated frames and ground truth frames, following~\cite{ruiz2022dreambooth}.

\topic{Implementation details.}
We implement the base text-to-video generative transformer model following Phenaki~\cite{villegas2022phenaki}.
Our implementation of the base model has 1.3B parameters, and our FACTOR model contains 1.6B parameters. The base model is trained for 1M steps, and our model is trained for 500K steps at a batch size of 256 and 128, respectively. The videos are generated at the resolution of 192$\times$320. During training, we use a mixture of 20\% images and 80\% videos in each batch. %

\topic{Compared methods.} 
Because our goal is to add fine-grained control to an existing text-to-video model, we evaluate whether our method can improve the generative quality of the base model for the fine-grained control inputs.
We validate that our approach enhances the base model's controllability without compromising the quality of the output video.
Direct comparison with other works (\eg, ~\cite{wu2022tuneavideo,chen2023controlavideo}), which inject dense structural control into the video generation model, is difficult due to their incompatibility with the sparse trajectory inputs our work utilizes.

To further show that our method provides reasonable controllability compared to existing text-to-video models, we present comparisons with state-of-the-art models when quantitative/qualitative results are available in their papers, given most of them are not publicly available.
We show the results of Imagen-Video~\cite{ho2022imagen}, MagicVideo~\cite{zhou2023magicvideo}, VideoLDM~\cite{blattmann2023videoldm}, Make-A-Video~\cite{singer2023makeavideo}, and ModelScope~\cite{wang2023modelscope}.
We evaluate two variants of our methods: 1)~\textbf{\ours-traj} ~denoted as our model with text and trajectory control. 2)~\textbf{\ours}~denoted as our full model with text, trajectory, and appearance control.
Please see the appendix for details.

\begin{figure*}[t]
    \vspace{-6pt}
    \centering
    \includegraphics[width=1.0\textwidth]{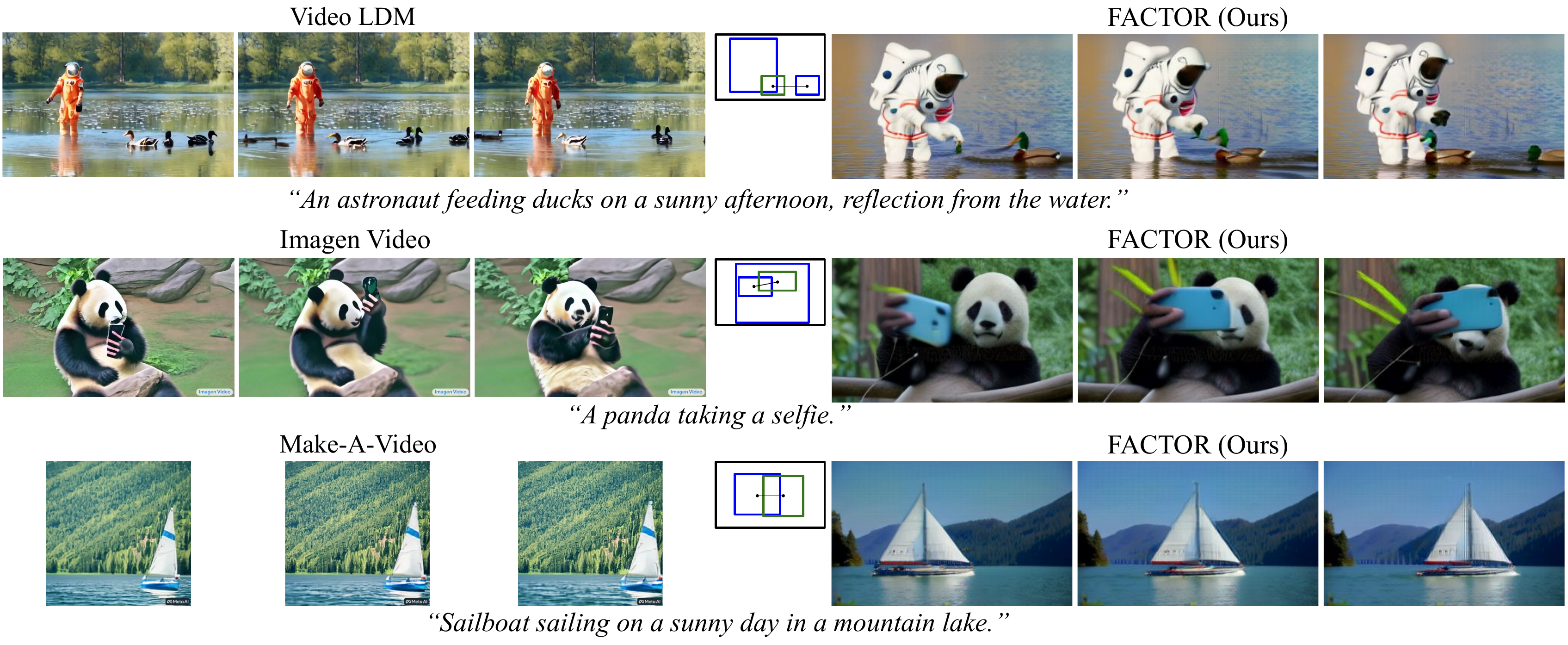}
    \vspace{-15pt}
    \caption{\textbf{Comparison to state-of-the-art.} FACTOR generates videos with better interaction between subject and objects. It shows more semantic meaning, \eg, the astronaut stretches out the hand, and the duck gets close to the astronaut, and better movement due to our fine-grained control.
    As the state-of-the-art models are not publicly available, we copy the results from their papers. 
    The blue and green boxes show the location of the object in the first and last frames, respectively.
    }
    \label{fig:comparison}
\end{figure*}

\begin{figure*}[t]
    \centering
    \includegraphics[width=1.0\textwidth]{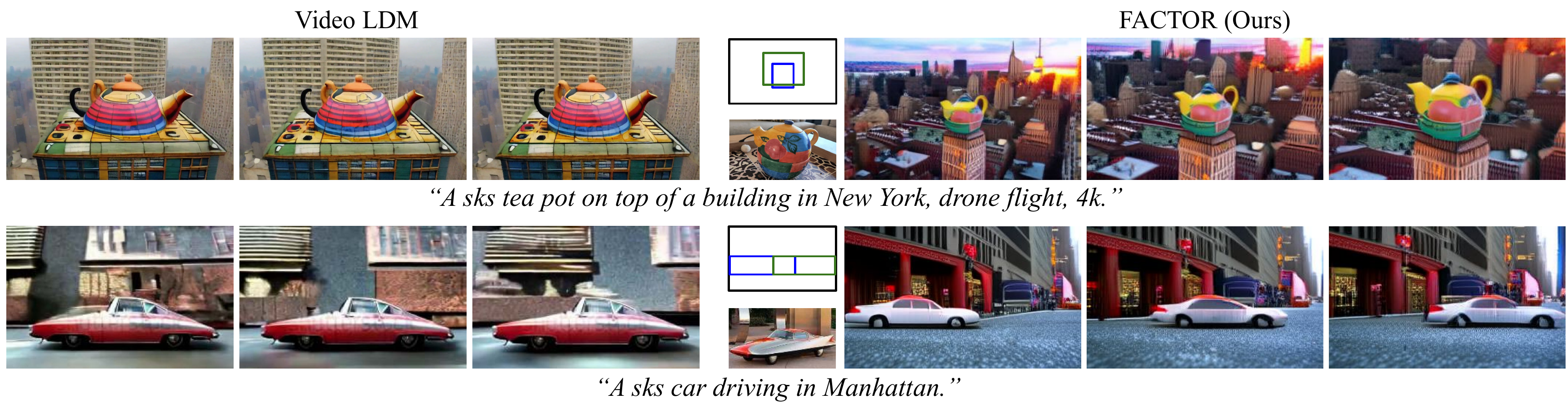}
    \vspace{-15pt}
    \caption{\textbf{Comparison of appearance control.} VideoLDM~\cite{blattmann2023videoldm} uses a backbone finetuned on multiple images while FACTOR is finetuning-free and conditions on a single reference image. FACTOR generates subjects that better align with the reference appearance, showing larger movements and better compatibility with the new environments. %
    }
    \label{fig:custom}
    \vspace{-6pt}
\end{figure*}

\subsection{Qualitative Results of Trajectory Control}
\label{sec:quanitative}

First, we assess the model's ability to control the trajectories of generated entities.
We design several prompts, including the description of trajectories such as the entity moving ``from the left to the right'' and the camera moving ``towards'' the entity.
In~\cref{fig:direction}, we show that given the trajectory prompts as input, Phenaki~\cite{villegas2022phenaki} fails to generate the correct movement direction for the elephant, and it fails to generate the lion with proper size and movement in the video.
In contrast, FACTOR-traj, which takes the hand-drawing trajectories as control inputs, generates the correct moving direction of the entity and the camera.

In~\cref{fig:motion}, we further present the results of trajectory control by drawing the trajectory of two main entities in the videos, such as \textit{(elephant, teddy bear)} in the top-left video and \textit{(monkey, motorcyclist)} in the bottom-left video.
By taking the trajectories as input, FACTOR-traj generates videos with two entities having the movement following the trajectories given by the users.
Moreover, FACTOR-traj brings an additional benefit of generating complex videos, including 1) interaction between subject-object such as \textit{``pick up"} and 2) interaction between subject-subject such as \textit{``high five with"}, even though our method is not specifically training on the examples of relation/interaction.

\subsection{Qualitative Results of Appearance Control}

Next, we present the results of appearance control, \ie, subject customization.
We select the reference images of the subjects from the datasets collected by DreamBooth~\cite{ruiz2022dreambooth}, CustonDiffusion~\cite{kumari2022customdiffusion} and VideoLDM~\cite{blattmann2023videoldm}.
For each subject, we use a single reference image as the appearance condition. 
\cref{fig:custom2} shows that FACTOR successfully generates the subject aligned with the reference appearance image. FACTOR handles interaction (milk and vase), the composition of two customized subjects (teapot and duck toy), and the motion of live subjects (cat walking, walking in the shoe).

We observe several limitations of our approaches. First, we find that the generated live subjects have limited motion in FACTOR compared to FACTOR-traj since our current framework conditions the subject appearance in the whole video uses a \textit{single} share image.
This could be further improved by applying data augmentation on the reference image in pixel or feature space~\cite{Khachatryan2023text2video} to generate more motion dynamics.
On the other hand, FACTOR has sub-optimal performance when the text prompt is not aligned with the trajectory/appearance control.
In this case, the model might perform inferior to the models that take only a single condition as input.

\subsection{Comparison to State-of-the-Art}
\label{sec:visual_comparison}
We further compare our video generation results with recent state-of-the-art text-to-video models. As their models are not publicly available, we present the results in their papers.
As shown in~\cref{fig:comparison}, by taking the additional trajectory control as input, FACTOR-traj synthesizes videos showing better interaction between subject and object, \eg, the astronaut stretches out the hand, the duck gets close to the astronaut, as well as better movement of the entity such as the movement of the boat.
In contrast, VideoLDM~\cite{blattmann2023videoldm} generates videos with less semantic meaning -- there is no action of feeding the duck, and Make-A-Video~\cite{singer2023makeavideo} produces videos with less movement.
We note that our method's ability is limited by the resolution/capability of the base model, and the method can be extended to other text-to-video models with better visual quality (\eg~\cite{ho2022imagen,blattmann2023videoldm}) if available.

To evaluate the ability of appearance control, we further compare FACTOR to VideoLDM~\cite{blattmann2023videoldm}, where they insert trained temporal layers into a DreamBooth backbone of Stable Diffusion model to generate video of customized subjects.
Their appearance control method requires multiple reference images and per-subject finetuning. On the other hand, FACTOR 1) does not require fine-tuning, and only an inference pass is needed, 2) requires a single reference image as input, and 3) achieves subject customization for multiple entities.
\cref{fig:custom} shows that our videos present better reconstruction of the customized subject. Using the additional trajectories as input, we show larger movements and better text alignment than VideoLDM, as their backbone is finetuned on static images.

\begin{table}[t]
    \small
    \centering
    \tabcolsep=0.05cm
    \caption{\textbf{Quantitative results on MSR-VTT.} Our model achieves a lower FVD compared to baselines. The high AP and CLIP-V scores show our model successfully generates videos aligned with the trajectory and appearance control inputs.
    }
    \vspace{-1mm}
    \label{tab:msrvtt}
    \begin{tabular}{lcccc}
        \toprule
        Methods & FVD~$(\downarrow)$ & CLIP-T~$(\uparrow)$ & AP~$(\uparrow)$ & CLIP-V~$(\uparrow)$\\
        \hline
        MagicVideo & 1290 & --- & --- & --- \\
        VideoLDM & --- & 0.2929 & --- & --- \\
        Make-A-Video & --- & 0.3049 & --- & --- \\
        ModelScope & 550 & 0.2930 & --- & ---\\
        \midrule
         Phenaki & 384 & 0.2870 & 0.0990 & 0.6626 \\
        \ours-traj~(Ours) & 317 & 0.2787 & 0.2902 & 0.6825 \\
        \ours~(Ours) & 124 & 0.2723 & 0.3407 & 0.7575 \\
        \bottomrule
        
    \end{tabular}
\end{table}

\subsection{Quantitative Results}
\topic{Results on MSR-VTT.}
As shown in \cref{tab:msrvtt}, FACTOR achieves better FVD scores than other text-to-video models.
In addition, FACTOR obtains a CLIP-T score comparable to state-of-the-art approaches. 
Our model has to generate videos that align with text and the fine-grained control signal including trajectory and reference appearance. 
Thus, it obtains a slightly worse CLIP-T score than state-of-the-art text-to-video models designed to optimize this metric. 

We further compare our implementation of Phenaki,~\ours-traj, and~\ours~as these models are built upon the same video backbone and training data while taking different levels of control as inputs, \ie, text only, text+trajectory, text+trajectory+appearance.
We find that with additional trajectory and reference images as input, the video quality score FVD improves significantly. %
We further compare the AP score. Although Phenaki is not trained to generate videos with desired object trajectories, we interpret its AP score as a chance performance when the text prompt is sufficient to generate videos that align with the user input trajectory.
~\ours-traj's higher AP score compared to Phenaki proves that the model successfully generates videos that align with the given trajectory control. 
On the other hand,~\ours~achieves better CLIP-V scores upon \ours-traj and Phenaki, showing that the model generates videos that better resemble the ground truth videos, indicating better alignment with the appearance control.

\begin{table}[t]
    \small
    \centering
    \tabcolsep=0.05cm
    \caption{\textbf{User study.} 
    We present the percentage of raters that prefer each method. \ours-traj only uses text and trajectory as control.
    Our model has a similar visual quality as the base model. The users prefer our results on text, trajectory, and appearance alignments to the ground truth videos. 
    }
    \label{tab:user}
    \begin{tabular}{lcccc}
        \toprule
         & Quality & Text & Trajectory & Appearance \\
         \hline
        Phenaki \vs \ours-traj & 38/62 & 27/73 & 18/82 & 20/80 \\
        Phenaki \vs \ours & 37/63 & 26/74 & 8/92 & 2/98 \\
        \ours-traj \vs \ours & 40/60 & 47/53 & 37/63 & 22/78 \\
        \bottomrule
    \end{tabular}
\end{table}

\topic{User study.}
To further augment the quantitative evaluation, we conduct a user preference study.
We present two videos generated by different methods using the same prompt and the ground truth video. We ask the raters to select their preferred results based on the following criteria: 
1) \textit{Quality}: which video has the better visual quality, 2) \textit{Text alignment}: which video is better aligned with the text prompt, 3) \textit{Trajectory alignment}: which video has the object trajectory that is better aligned with the ground truth video, and 4) \textit{Appearance alignment}: which video better reproduces the object and scene appearance of the ground truth video. 
We conduct the study with 30 pairs of videos and 5 participants. 

\cref{tab:user} show that FACTOR achieves better visual quality ($>$60\% preference) and text alignment ($>$70\% preference) compared to baselines.
By taking the additional trajectory and reference images as input, FACTOR's results have better trajectory/appearance alignment with the ground truth video, showing the effectiveness of our control module. 
Comparing the two variants of our model, FACTOR taking the reference images as a condition achieves better appearance alignment, presenting its capability of subject customization.

%% file: sec5_conclusions.tex
\section{Conclusions}
\label{sec:conclusions}
We present FACTOR, an approach to fine-grained controllable video generation. In addition to the text prompt, users can easily control the video generation process by naming the entities, drawing their trajectories, and providing their appearance through visual examples.
Our key idea is to adapt the text-to-video model to generate videos conditioned on fine-grained control by training an adaptive attention layer, and we model the interaction between inputs through a joint encoding module.
Our method generates videos with controllability on object appearance and trajectory. It brings additional benefits of generating complex videos including interaction.

%% file: sec6_appendix.tex
\section*{Appendix}

We present the implementation and experiment details in the appendix, including the details of datasets, training, model architectures, and evaluation metrics.

\topic{Datasets.}
We train the model on the WebVid-10M dataset. 
The dataset contains videos at 4 fps with a resolution of $256\times454$ and an average length of 40 frames.
During training, the videos are simply resized to our output resolution of $96\times160$ without data augmentation. We randomly sample 11-frame sub-sequences as training samples. 
To obtain our fine-grained annotations, we apply an off-the-shelf object detector~\cite{raid} to obtain the list of objects in the 40-frame videos. We run a tracking algorithm~\cite{Bewley2016_sort} to extract the trajectory of each object.
We sort the trajectories by the number of pixels they cover and remove the smaller trajectories that heavily overlap with the larger ones.
Finally, we select 12 object trajectories for each video in our annotations. At training time, the order of the objects is randomly shuffled.

To obtain the reference appearance images, we use the detected trajectories to crop the video frames. 
We randomly select one cropped region from the 40-frame videos for each object as the reference appearance image.
In addition to video, we use image data from the WebLI dataset in our training, which is used 20\% of the time in each batch. Similarly, we run an object detector~\cite{raid} to obtain the bounding box annotations. The image frames are repeated to the video length as a training sample.

\topic{Model architectures and training details.}
Our base text-to-video model contains a C-ViViT encoder-decoder and a bidirectional transformer.
The C-ViViT model is trained to compress the video with dimensions $11\times96\times160\times3$ into discrete tokens with dimensions of $6\times12\times20$.
The model has a hidden size of 512, an embedding dimension of 32, an MLP size of 2048, 8 heads, and a codebook size of 8192. It contains 4 layers of spatial transformers and 4 layers of temporal transformers. The model has 100M parameters.
The model is trained using the AdamW optimizer with a learning rate of $1\times10^{-4}$, a batch size of 128, and 1M iterations.

The bidirectional transformer is trained to produce videos at $11\times96\times160\times3$. 
The transformer takes a sequence length of 1920. It contains an embedding dimension of 1728, an MLP size of 4096, 36 heads, and 24 layers. The model has 1.3B parameters.
The model is trained using the AdamW optimizer with a learning rate of $4.5\times10^{-5}$, a batch size of 512, and 1M iterations.
The videos are generated at a unit of 11 frames and can be extended to an arbitrary length by applying a sliding window approach. Specifically, we first generate an 11-frame clip and use the last 5 frames as a condition for the model to generate another 6 frames. This process can be applied iteratively.
We apply a super-resolution model to increase the video resolution to $192\times320\times3$.
In our method, the C-ViViT model and the super-resolution model are kept as the original models for text-to-video generation, and only the bidirectional transformer model is trained to take the fine-grained control inputs.

As the video is compressed into a size of $6\times12\times20$, the control signals are given at 6 timesteps to generate 11 frames.
The control signal at each timestep contains prompt embeddings and control embeddings.
The length of the prompt embedding is 64, and the length of the control embedding is 220.
In FACTOR model, the control embedding is formed by the entity control of four objects, including the embedding of their description, location, and appearance with lengths of 1, 4, and 50, respectively.

To obtain the appearance embedding, the reference images are resized to $224\times224$ and input to a ViT-B/32 model to extract the appearance features. We use the grid features, which have a length of 50 and a size of 768.
The appearance embeddings are then input to an MLP layer to reduce its size to 512.
We use a learned embedding layer to obtain the entity description and location sequences with a feature size of 512.
The three contexts are then concatenated to form an embedding sequence of length 220 and input to a joint transformer encoder, which contains a self-attention layer to obtain the contextualized embeddings.
The embeddings are then split into prompt embeddings of length 64 and control embeddings of size 220 and input to the cross-attention layer and adaptive cross-attention layers separately.
In total, the FACTOR model contains 1.6B parameters.
At training time, the embedding layer, the joint encoder, and the adaptive cross-attention layers are learned while other layers are fixed.
The model is trained using the AdamW optimizer with a learning rate of $4.5\times10^{-5}$, a batch size of 128, and 500k iterations.
To apply classifier-free guidance, at training time, all the input conditions are dropped simultaneously 10\% of the time. At inference time, we use a classifier guidance scale of 12, an inference temperature of 4, and 48 sampling steps.

\topic{Evaluation metrics.}
We present the details of evaluation metrics.
1)~\textit{Fr\'echet Video Distance (FVD)} assesses the absolute video quality by measuring whether the distribution of generated videos is close to that of real videos in the feature space. We use the I3D model trained on Kinetics-400 for video features. 
2)~\textit{CLIP text similarity (CLIP-T)} assesses the alignment between input prompts and generated video frames using CLIP embeddings.
We use the ViT-B/32 model to compute the CLIP-T score. The score is reported as the average of all frames.
3)~\textit{Average Precision (AP)} assesses the alignment between input object trajectories and the generated videos. We apply a pre-trained object detector~\cite{raid} to compare the detected bounding boxes between the generated frames and ground truth frames and report the AP score. The AP score is computed at the $6^{th}$ frame.
4)~\textit{CLIP video similarity (CLIP-V)} assesses the similarity between the reference images and object appearance in the generated video frames. We measure the CLIP embedding distance between the generated frames and ground truth frames, following~\cite{ruiz2022dreambooth}.
Specifically, we use the bounding box coordinates in the trajectory of an object to crop the ground truth video frames. After obtaining a cropped image per frame containing that object, we randomly select one as the reference appearance image. Since the contents in the ground truth videos are used as references, the generated video should mostly reconstruct the ground truth video.
We measure the similarity of the generated and ground truth video frames and report the average.
We use the ViT-B/32 model to compute the CLIP-V score.